\documentclass[11pt,a4paper]{article}
\usepackage[hyperref]{acl2021}
\usepackage{times}
\usepackage{latexsym}

\usepackage{microtype}

\aclfinalcopy 

\usepackage{url}
\usepackage{amsmath}
\usepackage{mathrsfs} 

\usepackage{graphicx}
\usepackage{color,xcolor}
\usepackage{multirow}

\usepackage[utf8]{inputenc} 
\usepackage[T1]{fontenc}    
\usepackage{url}            
\usepackage{booktabs}       
\usepackage{amsfonts}       
\usepackage{nicefrac}       
\usepackage{microtype}      

\usepackage[ruled,vlined]{algorithm2e}

\title{Controlling Text Edition by Changing Answers of Specific Questions}
\author{Lei Sha,  Patrick Hohenecker, Thomas Lukasiewicz \\
  Department of Computer Science \\
  University of Oxford, United Kingdom \\
  {\tt \{lei.sha,  thomas.lukasiewicz\}@cs.ox.ac.uk} \\ \tt patrick@serein.ai}

\date{}

\begin{document}
\maketitle
\begin{abstract}
In this paper, we introduce the new task of \emph{controllable text edition}, in which we take as input a long text, a question, and a target answer, and the output is a minimally  modified text, so that it fits the target answer.
This task is very important in many situations, such as changing some conditions, consequences, or properties in a legal document, or changing some key information of an event in a news text.
This is very challenging, as it is hard to obtain a parallel corpus for training, and we need to first find all text positions that should be changed and then decide how to change them. 
We constructed the new dataset \textsc{WikiBioCTE} for this task based on the existing dataset \textsc{WikiBio} (originally created for table-to-text generation). We use \textsc{WikiBioCTE} for training, and manually labeled a test set for testing.  We also propose novel evaluation metrics  and a novel method for solving the new task. Experimental results on the test set show that our proposed method is a good fit for this novel NLP task.

\end{abstract}

\section{Introduction}
In many cases, we need to change some specific content in a document. For example, in the legal domain, the items and conditions in contract documents often need to be revised many times. We would like to use artificial intelligence to conduct this process for human editors.
A major difficulty of this process is that the machine learning model should decide where to edit and how to edit.

\begin{figure}[!t]
    \centering
    \!\!\!\includegraphics[width=1.1\linewidth]{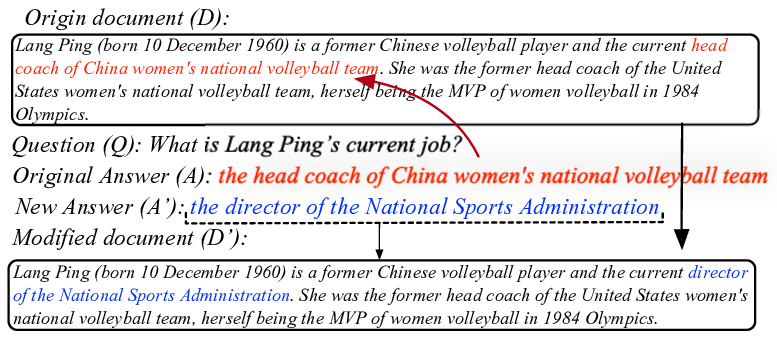}
    \caption{The original text $D$ is in the upper box. The question $Q$ to $D$ has an answer $A$ (in red; its rationale in $D$ also in red). If we would like to change the answer to the new answer~$A^\prime$ (in blue), then we have to change some content in~$D$, yielding the modified text $D'$ (with the new content in blue) in the lower box.}
    \label{fig:exp}
\end{figure}

Usually, the place of specific content (``where to edit'') can be located by a question, and the content updating (``how to edit'') can be determined by the answer of the question. 
Therefore, in this paper, we propose the new task of \emph{controllable text edition (CTE)}. In this task, we would like to achieve the following goal: \textit{adjust some content of a document $D$, to make the answer $A$ of a document-related question~$Q$ changed to a new answer $A'$}. For example, in Fig.~\ref{fig:exp}, when we change the red part of the original text to the blue part, the answer of the question turned to the new answer as a consequence.  

There are three main challenges in this task:

\smallskip 
\noindent (1) The machine learning model should decide the positions that need to be changed in the document. Usually, finding the answer positions for a given document-related question is similar to extractive machine reading comprehension tasks~\cite{zeng2020survey}, which requires to fully understand both the question and the document. Nearly all extractive machine reading tasks, such as SQuAD~\cite{rajpurkar2016squad,rajpurkar2018know} and CNN/Daily Mail~\cite{hermann2015teaching}, focus on extracting one span from the document as answer.
Differently from extractive machine reading, in our task, the answer $A$ is not necessarily a substring of the document, and there may exist multiple positions that have to be changed. Therefore, our task is much more challenging than extractive machine reading.

\smallskip 
\noindent (2)  The model should generate a new document that supports the new answer $A^\prime$ for question $Q$. Note that this cannot be solved by directly replacing the original words in the edit positions with the new answer $A^\prime$, because the new answer may not fit perfectly with the document, which would make the document disfluent. 

\smallskip 
\noindent (3) There are nearly no parallel data for model training, because obtaining a large annotation set for this task is very hard.\footnote{To annotate a large parallel dataset, we need to prepare a document, a document-related question, and its expected answer. Then, the data grader should provide an adjusted version of the document that satisfies the expected answer, which requires the data grader to have a high education level.} However, the model may be trained by  lists of triples $\langle Q, D, A\rangle$ that can be  obtained from datasets in machine reading and/or structured data extraction (as described below). 

\smallskip 
In this paper, we introduce and define the task of controllable text edition (CTE). We propose to 
transform the 
\textsc{Wikibio} dataset~\cite{lebret2016neural} into a list of triples $\langle Q, D, A\rangle$ for training. \textsc{Wikibio} was originally designed for table-to-text generation, in which each case is composed of a Wikipedia passage $D$ and an infobox (which is a list of $\langle$field, content$\rangle$\footnote{In the Wikipedia Infobox, ``field'' represents the type of information (such as \textit{Name}, \textit{BirthDate}, and \textit{Known for}), while ``content'' represents the value of ``field''. } pairs). In detail, we take each ``field'' in the infobox as the question $Q$, and each ``content'' in the infobox as the answer $A$. Therefore,  for each $\langle$field, content$\rangle$ pair, we can create a  $\langle Q, D, A\rangle$ triple. After some pruning, we finally selected $26$ different $Q$'s and $141k$  $\langle Q, D, A\rangle$ triples for the training set, as well as $17.7k$ triples for the development set. We also annotated a small test set of about $1k$ data for evaluation in the form of $\langle Q, D, A, A', D'\rangle$ ($A'$ represents the new answer, and $D'$ represents the ground-truth modified text). The resulting new  dataset is called \textsc{WikiBioCTE}. 

In addition, we propose a novel method, called \textit{Select-Mask-Generate (SMG)}, to solve the proposed CTE task. In this method, we use the se\-lec\-tor-predictor architecture by  \newcite{sha2021learn} to select the answer-related tokens, and we then use complementary masks to split the text into an answer-related part and an answer-unrelated part. Then, we reconstruct the original text based on   the answer-unrelated part and the original answer. The reconstruction process is a partial generation method, which only generates the masked-out part without any length limit. 
In our experiments, the SMG model has achieved the state-of-the-art performance, compared to  baseline models in the generation of modified documents. The code and the  test set \textsc{WikiBioCTE} are available online\footnote{https://sites.google.com/view/control-text-edition/home}.

\section{Related Work}
The proposed task of \textit{controllable text edition} is related to the following existing tasks.

\subsection{Attribute Disentanglement }
Attribute disentanglement tends to control the attributes of a given text or image (such as sentiment, tense, syntax, or face pose) by disentangling different attributes into different subspaces. When transferring attributes, the content of the text/image needs to be preserved. Usually, disentanglement works can be divided into implicit and explicit disentanglement. Implicit disentanglement~\cite{higgins2017beta,chen2018isolating,moyer2018invariant,mathieu2018disentangling,kim2018disentangling} separates the latent space into several components in a purely unsupervised way, expecting that each component corresponds to an attribute. However, the number of components cannot be decided in advance, neither does the correspondence between attributes and components. Also, the training process may prune some of the components~\cite{stuhmer2019independent}, which will hurt the interpretability of the latent space. Explicit disentanglement~\cite{chen2016infogan,john-etal-2019-disentangled,romanov2019adversarial,sha2021multi} tends to separate the latent space into more interpretable components with explicit correspondence to specific attributes. Hence, it usually requires gold labels of attributes in the training set.  

In comparison, our task tends to control the content of the text by tuning answers to text-related questions. Attribute disentanglement is difficult to be applied to our task, because the modification of the content should be decided by both the question and the answer simultaneously, which is much sparser than attributes.

\subsection{Lexically Constrained Decoding}
Lexically constrained decoding~\cite{hokamp-liu-2017-lexically,miao2019cgmh,sha-2020-gradient} directly controls the output of the generation model by adding constraints. Usually, the constraints include hard constraints (requiring the generated sequence to contain some keywords) and soft constraints (requiring the generated sentence to have the same meaning to a given text). The basic methods of  lexically constrained decoding can be divided into enhanced beam search~\cite{hokamp-liu-2017-lexically,post-vilar-2018-fast} and stochastic search~\cite{miao2019cgmh,liu-etal-2020-unsupervised,sha-2020-gradient}. Enhanced beam search~\cite{hokamp-liu-2017-lexically,hasler-etal-2018-neural,hu-etal-2019-improved} changes some strategies in beam search to make the process of searching for a constraint-satisfying sentence  easier. However, for some tasks with an extremely large search space, beam-search-based methods may be computationally too costly or even fail~\cite{miao2019cgmh}. Stochastic search tends to edit an initial sentence step-by-step, where the editing position and action can be decided by Metropolis-Hastings sampling~\cite{miao2019cgmh}, a discrete scoring function~\cite{liu-etal-2020-unsupervised}, or gradient-based methods~\cite{sha-2020-gradient}.  However, lexically constrained decoding is hard to be applied to our task, because adjusting the text to fit a text-related question's new answer is much more complicated than simply satisfying a hard or soft constraint.
\begin{table}[!t]
	\textbf{Table:}\\[.2cm]
	\resizebox{\linewidth}{!}{
		\footnotesize
		\begin{tabular}{r|lp{5cm}|}
			\cline{2-3}
			\textbf{ID} & \textbf{Field} & \textbf{Content}\\
			\cline{2-3}
			1\!\! &\!\! Name  \!\!&\!\! \textit{Frank Fenner}\\
			2\!\! &\!\! Born  \!\!&\!\! \textit{21 December 1914, Ballarat}\\
			3\!\! &\!\! Died  \!\!&\!\! \textit{22 November 2010 (aged 95) Canberra}\\
			4\!\! &\!\! Occupation \!\!&\!\! \textit{Virology}\\
			5\!\! &\!\! Nationality \!\!&\!\! \textit{Australian}\\
			6\!\! &\!\! Known for \!\!&\!\! \textit{Eradication of smallpox, Control of Australia's rabbit plague}\\
			\cline{2-3}
		\end{tabular}
	}\\
	
	\textbf{Text:}
	{\small Frank John Fenner (21 December 1914 – 22 November 2010) was an Australian scientist with a distinguished career in the field of virology. His two greatest achievements are cited as overseeing the eradication of smallpox, and the control of Australia's rabbit plague by introducing the Myxoma virus.}
	
	\caption{An example of a Wikipedia infobox and a reference text.}\label{tab:example}
\end{table}
\subsection{Text Editing and Infilling}

In some tasks, to simplify the text generation problem, researchers tend to edit existing text or prototypes to obtain a refined text that satisfies some specific requirements. Examples are the generation of summaries by template-based rewriting~\cite{cao-etal-2018-retrieve,hashimoto2018retrieve} 
and the generation of text or a response by editing a prototype sentence~\cite{guu-etal-2018-generating,pandey2018exemplar,wu2019response}.  In \cite{yin2018learning}, the distributed representations of edit actions are learned and applied  to editing Wikipedia  records~\cite{faruqui-etal-2018-wikiatomicedits} and Github code~\cite{yin2018learning}. \newcite{panthaplackel2020copy} further integrate a copy~mechanism into text editing. 

Text infilling~\cite{fedus2018maskgan} means to use machine learning models to fill the blanks of a cloze test.
\newcite{zhu2019text} propose a more general text infilling task, which allows an arbitrary number of tokens (instead of a single token) in each blank.

In the above text editing tasks, the goal of editing is always consistent among all the datasets: for a better summarization, a better response, or a better informative sentence. Differently from them,  our proposed task requires the editing to be guided by the document-related answer of the question. So, each above case has a different editing goal. Thus, our task requires deciding where to edit according to the given question in the first step, and then deciding how to edit, which makes our task more complicated than all the above text editing tasks.

\section{Dataset}
 We now formally define the task of controllable text edition
 and propose a dataset for this task. 

\subsection{Task Definition}
The task of \emph{controllable text edition (CTE)} is defined as follows. The input is a triple $\langle D, Q, A^\prime\rangle$, where $D$ is a document, $Q$ is a document-related question, and $A^\prime$ is an expected answer for $Q$ to $D$. The output is $D^\prime$, which is a minimal modification of $D$ such that the answer for $Q$ to $D'$ is now  $A^\prime$.
Note that the original answer of $Q$ to $D$ is $A$, but $A$ is not an input to the task, and usually $A\ne A^\prime$. 

\subsection{\textsc{WikiBio} as Controllable Text Editing Dataset}
We propose to modify the \textsc{WikiBio} dataset~\cite{lebret2016neural} to make it fit for our task. \textsc{WikiBio} was originally designed for table-to-text generation~\cite{lebret2016neural,sha2018order,liu2018table}, which generates a celebrity's biography according to his/her basic information. Each example in the dataset is composed of a Wikipedia infobox and a text~(the first paragraph in the Wiki page)  describing the infobox as shown in Table~\ref{tab:example}.

In an inverse way, the \textsc{WikiBio} dataset can be taken as a question-answering dataset: each \textit{field} can be taken as a question, and each \textit{content} can be taken as an answer. For example, in Fig.~\ref{tab:example}, the \textit{field} ``Occupation'' can be interpreted as question ``What is the person's occupation?'', and the corresponding \textit{content} ``Virology'' is the answer. 

Therefore, we  take the \textit{text} in \textsc{Wikibio} as the document ($D$) in our task, the \textit{field} as the question~($Q$), and the \textit{content} as the answer ($A$). Due to the huge cost of data annotation, the model needs to be trained without the changed answer ($A^\prime$) and the referenced document ($D^\prime$). 

For the creation of the training and development sets, we count the frequency of \textit{fields} and select the \textit{fields} that occurred more than $5$k times in \textsc{Wikibio}'s training set as candidate questions ($Q$'s). Then, we filter out some $Q$'s that do not have corresponding answers in $D$\footnote{Since $D$ is the first paragraph in the Wikipedia page, it~usually does not contain everything mentioned in the infobox, such as \textit{death cause} and \textit{high school}.}. We then get a list of $26$ different $Q$'s as  shown in Table~\ref{tab:selected_fields}.
After filtering the~$Q$'s according to Table~\ref{tab:selected_fields},  we get $141k$  $\langle Q, D, A\rangle$ triples for the training set and $17.7k$ triples for the development set.

Then, we manually labeled a small test set in which each example contains $(D, Q, A)$ as well as the changed answer ($A^\prime$) and the referenced document ($D^\prime$). 
The annotation process can be illustrated as follows: 
\begin{enumerate}
\item We randomly sampled an equal number  of examples  for all the \textit{fields} in Table~\ref{tab:selected_fields}. For each \textit{field}, we sample $\lceil\frac{1000}{\#F}\rceil$ cases ($\#F$ is the number of selected \textit{fields}), to make sure that the size of the test set is around $1$k.

\item We assigned a changed answer ($A^\prime$) to each example by randomly picking a similar phrase to the original answer ($A$). The similar phrase may occur in different examples, but it shares the same $Q$ with the original answer ($A$).

\item We asked human data graders to give a modified text ($D^\prime$) for each example according to the original text ($D$), question ($Q$), and the changed answer ($A^\prime$). We asked two talented linguistics to annotate the $1$k  test set. 
\end{enumerate}

Note that there are also other datasets that are potentially able to be modified as controllable text editing dataset, such as SQuAD~\cite{rajpurkar2016squad}, RACE~\cite{lai-etal-2017-race}, and  MCTest~\cite{richardson-etal-2013-mctest}. We did not choose them for the following reasons: (1) For extractive machine reading tasks like SQuAD~\cite{rajpurkar2016squad}, the answers are simple substrings of the document, so that in most cases, the text modification in our task can be solved by a simple string replacement, which violates the goal of our task. (2) Multiple-choice machine reading tasks like RACE~\cite{lai-etal-2017-race} usually require full and deep reasoning of the whole document to get the answer, which would make the text modification in our task unable to be solved by partial modification. Differently from them, most  \textit{contents} ($A$) in \textsc{WikiBio}  usually cannot be directly extracted as substrings from the document ($D$). Besides, the \textit{contents} usually has some related information that should be modified at the same time. For example, if somebody is a pianist, then he/she may have received a piano award instead of a guitar award. Therefore, \textsc{WikiBio}  satisfies the goal of our proposed task: making minimal changes to the original document to make it fit the changed answer ($A^\prime$).

\begin{table}[!t]
    \centering
    \resizebox{\linewidth}{!}{
    \begin{tabular}{cccccc}
    \toprule[1.0pt]
         birth date	&679.4k & name&675.8k&	birth place	&659.3k\\
         death date	&420.7k & death place&377.7k&occupation & 231.9k\\ 
         position & 199.4k & nationality&187.1k &	spouse&184.0k\\	
         fullname & 180.2k & alma mater&115.5k & children & 114.9k\\
         residence&112.1k & 	religion & 99.3k & predecessor&91.0k\\	successor&90.1k&	known for & 63.4k &origin & 46.6k\\
country&43.5k&	education&43.1k &	instrument & 36.7k	\\
college	&35.9k&citizenship&29.1k&ethnicity & 28.7k \\
discipline&11.2k & work institutions&	5.3k & &	\\
    \bottomrule[1.0pt]
    \end{tabular}
    }
    \caption{The selected \textit{fields} from \textsc{WikiBio} and their occurrence in \textsc{WikiBio}'s training set. These are taken as the questions ($Q$'s) in our proposed task.}
    \label{tab:selected_fields}
\end{table}

\section{Select-Mask-Generate (SMG) Method for Controllable Text Edition}


We introduce the training and testing method of our proposed method. In the training phase, the model is trained to learn to recognize answer-related ($A$-related) tokens and  learn to fill new-answer-related ($A'$-related) tokens into the blanks after deleting answer-related tokens.

\subsection{Training Phase}
In the training phase, we only have $Q$, $D$, and $A$. So, we teach the model to (1) identify answer-related information, and (2) be able to reconstruct $D$ from $A$ and ($D-A_p$) (the original text with all answer-related information masked out, where $A_p$ means the predicted answer-related tokens).

\begin{figure}[!t]
    \centering
    \includegraphics[width=\linewidth]{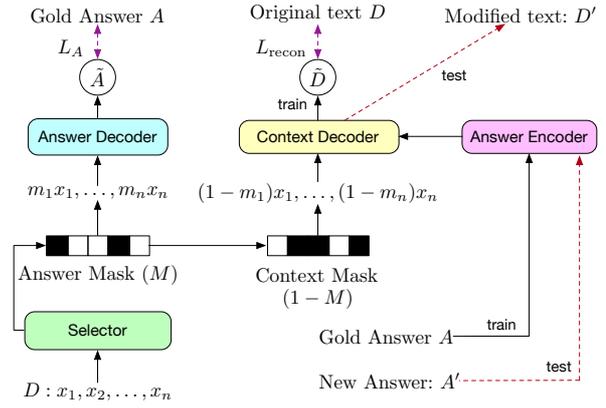}
    \caption{The architecture of our SME model. In the testing phase, we need to replace the input to the answer encoder from the gold answer $A$ to the new answer $A'$, then the output of the context decoder will become the modified text $\tilde D'$.}
    \label{fig:arch}
\end{figure}

The model architecture is shown in Fig.~\ref{fig:arch}.
Inspired by InfoCal~\cite{sha2021learn}, we use a \textit{Selector-Predictor} architecture to identify the least-but-enough answer-related words in the original document ($D$). The main architecture of the \textit{Selector} network is a BiLSTM model, which samples\footnote{The sampling process is implemented by Gumbel Softmax~\cite{jang2016categorical}, which is differentiable.} a binary-valued mask ($M$) for each input token (called answer mask), denoting whether to select this token as answer-related token ($1$) or not~($0$).
Given an input document $D=\{x_1,\ldots,x_n\}$ and a question $Q$, the \textit{Selector} samples an answer-related mask $M=\{m_1,\ldots,m_n\}$ as follows:
\begin{equation}
    M\sim\text{Sel}(M|D, Q),
\end{equation}
where ``Sel''  represents the selector network. Then, we call the complement  of the answer mask  ($\overline M = 1-M$) as the context mask, and we denote \textit{context  template} as  the token sequence after masking out the answer-related tokens.

\subsubsection{Answer Reconstruction}
We require that the answer-related information contains everything about the answer $A$, so we use an answer decoder to reconstruct an answer sequence~$\tilde A$. Then, we calculate the reconstruction loss $L_A$ as follows:
\begin{align}
    &p_a(\tilde A|M,D) = \text{Dec}_A(\frac{1}{\sum_j m_j}\sum_im_ix_i),\\
    &L_A = \mathbb E_{M\sim \text{Sel}(M|D,Q)} p_a(A|M,D),
\end{align}
where $\text{Dec}_A$ is the answer decoder, and $p_a$ is the sentence distribution generated by $\text{Dec}_A$. Note that the input to $\text{Dec}_A$ is the average vector of the selected token vectors: the answer-related tokens are usually very few, so it is not necessary to use heavier encoders like LSTMs~\cite{hochreiter1997long} or transformers~\cite{vaswani2017attention}.

\subsubsection{Document Reconstruction}
On the other hand, $D$ should be reconstructed by the \textit{context  template} and the gold answer $A$. We use an LSTM encoder $\text{Enc}_D$ to encode the \textit{context tokens} as shown in Eqs.~\ref{eq:hm1} and~\ref{eq:hm}:
\begin{align}
&h'_1,\ldots,h'_n = \text{Enc}_D([\overline m_1x_1,\ldots, \overline m_nx_n]),\label{eq:hm1}\\
&H_m=\text{Maxpooling}(h'_1,\ldots,h'_n),\label{eq:hm}
\end{align}
where $h'_1,\ldots,h'_n$ are the encoding vectors corresponding to each input token.
We then take the averaged word vector of the input gold answer $A$, denoted $V_A$, as an external condition of the decoder.

Differently from conventional decoders, our decoder only partially generates tokens to fill in the blanks of the \textit{context  templates}, as  shown in Fig.~\ref{fig:decoder}. This brings two changes in the training phase: (1) we only need to calculate the loss caused by the tokens filled in the blank, and (2) the model needs to learn an external end-of-answer (EOA) token $S_{eoa}$ for each token filled in the blanks.  The EOA token is very important because it is an indicator about when to stop filling the current blank. 

\paragraph{Learning to generate the words.} 
In each time step $t$ of the decoder, we use an LSTM~\cite{hochreiter1997long} unit to predict the next word $y_t$ and the EOA token $S_{eoa}$ as follows:
\begin{align}
&h_t = \mathcal F_\text{LSTM}([y_{t-1}, V_A], h_{t-1}),\\
&\begin{bmatrix} h_w \\ h_\text{eoa} \end{bmatrix}=
\begin{bmatrix} \sigma \\ \sigma  \end{bmatrix}\mathcal F_{m}(h_t),\\
&s^\text{lstm}_t(w) = \mathcal F_{w}(h_w),\\
&p(\tilde S_{eoa}(t)) = \text{Softmax}\big(\mathcal F_{eoa}( h_\text{eoa})\big),\label{eq:eoa}
\end{align}
where $h_w$ and $h_\text{eoa}$ are hidden layers (the time step index $t$ is omitted),  $\mathcal F_\text{LSTM}$ is an LSTM cell, $\mathcal F_{m}$, $\mathcal F_{w}$, and $\mathcal F_{eoa}$ are linear layers, and $s^\text{lstm}_t(w)$ is a scoring function that suggests the next word to generate. $p(\tilde S_{eoa}(t))$ is the probability distribution of the EOA token.

Note that in the decoder, we use the copy mechanism~\cite{gu-etal-2016-incorporating}, which encourages the decoder to generate words by directly copying from the input context sequence $D$ and answer sequence~$A$. The copy mechanism computes a copy score $s^\text{copy}_t(w)$ for each word in $D$ and $A$. Then, the generated probability of each word is computed~as:
\begin{align}
s_t(w) &= s^\text{lstm}_t(w) + s^\text{copy}_t(w),\\
p_t(w)&=\text{Softmax}(s_t(w)).\label{eq:pt}
\end{align}


Thus, the document $D$'s reconstruction loss is  as follows:
\begin{align}
    L_\text{recon} &= -\mathbb E_{M}\Big[ \sum_tm_t\log p_t(y_t|\overline M,A)\Big],
\end{align}
where $M\sim \text{Sel}(M|D,Q)$, the mask $m_t$ is multiplied in each time step, because we only need the losses of blank-filling tokens.

\paragraph{Learning the end-of-answer~(EOA) tags.} 
We have an EOA tag for each blank-filling token. 
The EOA tag is $1$  if the corresponding token is the last token in the blank. For the other  blank-filling tokens, the EOA tag is $0$. The gold EOA tag in each time step $g^\text{eoa}_t$ can be computed by the difference between the previous answer mask  $m_{t-1}$ and the current answer mask  $m_t$. There are three possible values ($-1$, $0$, and $1$): $g^\text{eoa}_t=0$ when the difference is $-1$ or $0$, and $g^\text{eoa}_t=1$ when the difference is $1$. Then, we have the cross-entropy loss as Eq.~\ref{eq:eoaloss}:
\begin{equation}\label{eq:eoaloss}
\small
\begin{aligned}
    g^\text{eoa}_t &= \max(m_{t-1}-m_{t},0)\\
    L_\text{eoa} &= -\mathbb E_{M}\Big[ \sum_t\Big(g^\text{eoa}_tm_t\log p(S_{eoa}(t)=1)\\
   &+(1-g^\text{eoa}_t)m_t\log p(S_{eoa}(t)=0)\Big)\Big].
\end{aligned}
\end{equation}
Therefore, the final  optimization objective is shown in Eq.~\ref{eq:L}:
\begin{equation}\label{eq:L}
    L=L_A + \lambda_rL_\text{recon} + \lambda_\text{eoa}L_\text{eoa},
\end{equation}
where $\lambda_r$ and $\lambda_\text{eoa}$ are hyperparameters.

\subsection{Inference Phase}
In the inference phase, we take the new answer $A^\prime$ as the input to the context decoder instead of the gold answer $A$. Then, the output of the context decoder will become the modified text $\tilde D^\prime$.

We choose an autoregressive partial generation method for inference. Our partial generation method can fill the blanks with any-length phrases and can directly replace any decoder, which cannot be done by any existing alternative methods. For example, in the method using global context~\cite{donahue-etal-2020-enabling}, it is an pretrained language model by itself. However, in our architecture, the masks are decided by the selector module. Therefore, even the number and length of the blanks cannot be decided before training. So, the ground-truth target sequence for the finetuning of the pretrained language model would also be hard to decide. Therefore, the partial generation method is the best choice for our task.

\subsubsection{Partial Generation}
\begin{figure}
    \centering
    \includegraphics[width=\linewidth]{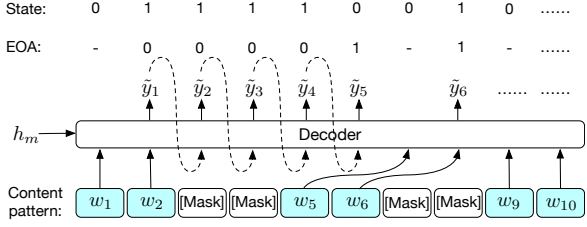}
    \caption{The partial decoding process. This process requires two tags (state and EOA tag) for indicating when to start generation and when to stop generation.}
    \label{fig:decoder}
\end{figure}

Since we already have a \textit{context  template}  when we are generating the modified document, we only need to generate tokens to fill the blanks in the \textit{context  template}. The partial decoding process is shown in Fig.~\ref{fig:decoder}. We use an indicator \textit{state}$=0$ to denote the \textit{reading} mode (reading the \textit{context  template} words), and  \textit{state}$=1$ to denote the \textit{writing} mode (generating the blank-filling words). The basic generating process is described as follows: when the model is \textit{reading} the  \textit{context  template}, if it meets a masked token, the mode turns to \textit{writing} mode, and it starts to generate words to fill the current blank. When the EOA tag turns to $1$, or the decoding length $l_g$ surpassed a limit $l_\text{max}$, the mode turns back to \textit{reading} mode. Note that this decoding process can generate an arbitrary number of words for each blank, and we can fill all blanks in a  \textit{context  template} in a single decoding pass, which is much more efficient than MaskGAN~\cite{fedus2018maskgan} and  text filler~\cite{zhu2019text}. The detailed algorithm is shown in Algorithm~\ref{algo:1}.

\begin{algorithm}[!t]\small
\SetAlgoLined
\KwIn{Context  template: $C$}
\KwOut{Generated Sequence: $\tilde D^\prime$}
\KwData{Read-write state: $S$, End-of-answer label: $S_{eoa}$, Context  template index: $I_c$, Local generate length: $l_g$, current input token $x_{in}$}
 $S\leftarrow 0$, $I_c\leftarrow 0$, $l_g\leftarrow 0$, $\tilde D^\prime \leftarrow  []$\; 
 Set the first input token $x_{in}\leftarrow C[0]$\;
 \For{each time step $t\leftarrow 1,2,\ldots$ }{
  Calculate $\tilde y_t$ by Eqn.~\ref{eq:pt}\;
  Calculate $S_{eoa}$ by Eqn.~\ref{eq:eoa}\;
  \If{$S=0$}{
  $\tilde D^\prime\leftarrow\tilde D^\prime + [C[I_c]]$\;
    \If{$C[I_c]\neq$`[M]' and $C[I_c+1]=$`[M]'}{
    $I_c\leftarrow I_c+1$\;
        \While{$C[I_c]=$`[M]'}{
        $I_c\leftarrow I_c+1$\;
        }
         \If{$S_{eoa}\neq 1$}{
            $S\leftarrow 1$\;
         }
    }
    \ElseIf{$C[I_c]\neq$`[M]' and $C[I_c+1]\neq$`[M]'}{
     $I_c\leftarrow I_c+1$\;
    }
  }
  \ElseIf{$S=1$}{
  $\tilde D^\prime\leftarrow\tilde D^\prime + [\tilde y_t]$, $l_g\leftarrow l_g+1$\;
         \If{$S_{eoa}=1$ or $l_g\ge l_\text{max}$}{
            $S\leftarrow 0$, $l_g\leftarrow 0$\;
         }
         }
  \If{$S=0$}{
      $x_{in}\leftarrow C[I_c]$\;
  }
  \ElseIf{$S=1$}{
      $x_{in}\leftarrow \tilde y_t$\;
  }   
  }
  \Return $\tilde D^\prime$\;
 \caption{The decoding process.}
\label{algo:1}
\end{algorithm}

\section{Experiments}
In the experiment part, we proposed some specific evaluation metric for our \textit{controllable text edition} task and then compare and analysis the performance of our proposed method (SMG) on the \textsc{WikiBioCTE} dataset.
\subsection{Evaluation Metrics}

For the evaluation of the modified document $\tilde D^\prime$,  we use the following two automatic evaluation metrics:

\smallskip 
\noindent (1) BLEU ($\tilde D^\prime$ vs.~$D^\prime$): This metric measures the BLEU score~\cite{papineni-etal-2002-bleu} between the generated modified document $\tilde D^\prime$ and the reference document $D^\prime$. 

\smallskip 
\noindent (2) iBLEU~\cite{sun-zhou-2012-joint}: This metric is previously widely used in evaluating paraphrase generation tasks~\cite{liu-etal-2020-unsupervised,sha-2020-gradient}. iBLEU is defined as:  $\text{iBLEU}=\text{BLEU}(\tilde D^\prime,D^\prime) - \alpha\text{BLEU}(\tilde D^\prime,D)$\footnote{$\alpha$ is set to $0.9$, which is consistent with previous works~\cite{liu-etal-2020-unsupervised}.}, which penalizes the similarity between the modified document $\tilde D^\prime$ and the original document $D$. The goal of this metric is to measure the extent to which the model directly copies words from the original document $D$ without taking any content from $A^\prime$.

\smallskip 
\noindent (3) diff-BLEU ratio: diff-BLEU is a BLEU score computed between $\tilde D^\prime$ and a \textit{difference sequence} between the gold modified document $D'$ and the original document $D$. The  \textit{difference sequence} is obtained by masking out the longest common sequence between $D$ and $D'$ from $D'$. Since this maximum value of this BLEU score is the BLEU value between the gold modified document $D'$ and the \textit{difference sequence}, we use their quotient as the diff-BLEU ratio score as shown in Eq.~\ref{eq:diff}:
\begin{equation}\label{eq:diff}
    \text{diff-BLEU ratio}=\frac{\text{BLEU}(\tilde D^\prime, D'-D)}{\text{BLEU}(D^\prime, D'-D)}.
\end{equation}

\smallskip 
\noindent (4) Perplexity: This metric measures the fluency of the generated content-modified document $\tilde D^\prime$. We applied a third-party language model (Kneser-Ney language model (\citeyear{kneser1995improved}))  as the perplexity evaluator. We trained the language model on the whole training set of \textsc{WikiBio}, and use the trained model as the evaluation of fluency, where a lower perplexity value is better.

\smallskip 
Besides, we used human effort to  evaluate two aspects of the content-modified document $\tilde D^\prime$. \textit{Correctness} is an accuracy score from $0.0\%\sim 100.0\%$, which  evaluates whether $\tilde D^\prime$ has successfully turned the answer of question $Q$ from $A$ to $A^\prime$. \textit{Fluency} is from $0.0\,{\sim}\, 5.0$, which  evaluates whether  $\tilde D^\prime$ is fluent from a human being's view. The scoring details are in the supplemental materials.

\begin{table}
    \centering
    \resizebox{\linewidth}{!}{
    \begin{tabular}{lccc}
    \toprule[1.0pt]
         & Seq2Seq & SMG (g) & SMG (p) \\
\midrule[0.5pt]
     BLEU ($\tilde D$ vs. $D$)    & 82.21 &\textbf{89.29} &87.53 \\
     iBLEU   & 5.63 & \textbf{10.05} &8.94\\ 
     diff-BLEU ratio&21.3\% &\textbf{62.5\%} &59.8\%\\
     Perplexity & \textbf{198}&235 &373\\
     Human (Correctness) &73.5\% &\textbf{80.2\%} &76.9\%\\
     Human (Fluency) &\textbf{4.56} &4.54 &4.32\\
     \bottomrule[1.0pt]
     \end{tabular}
     }
    \caption{The overall performance of all competing methods. SMG (g) denotes that the method SMG is using the gold templates for partial generation, and SMG~(p) denotes that the method SMG is using the predicted templates for partial generation.}
    \label{tab:overall}
\end{table}
     \begin{table}
    \centering
    \resizebox{\linewidth}{!}{
    \begin{tabular}{lccc}
    \toprule[1.0pt]
         &Random & Seq2Seq & SMG \\
\midrule[0.5pt]
     BLEU (predicted  template)&21.5 &59.5 &\textbf{89.1}\\
     Answer $F_1$ &0.14 &0.55 &\textbf{0.68}\\
    \bottomrule[1.0pt]
         \end{tabular}
         }
    \caption{Performance of answer-related words selection.}
    \label{tab:p}
\end{table}

\begin{table*}[!ht]
\renewcommand{\arraystretch}{1.4}
    \centering
    \resizebox{\linewidth}{!}{
    \begin{tabular}{|p{2cm}|p{5cm}|p{9cm}|p{10.3cm}|}
    \hline
    \multirow{8}{*}{Input}& $D$: george evans -lrb- born 13 december 1994 -rrb- is an \textbf{ english } footballer who plays as a \underline{ midfielder} \textbf{\underline{or   centre-back}  for} manchester city . & 
    $D$: andrei UNK -lrb- born 1975 in satu mare , romania  -rrb-  is  a retired romanian \textbf{\underline{aerobic gymnast}} . he had a successful career winning four world championships medals -lrb- two gold , one silver , and one bronze -rrb- after his retirement in 1997 he went with to germany where he works as a \underline{gymnastics} coach at the UNK \underline{gymnastics} club in hanover . & 
    $D$: andrew justin stewart coats -lrb- born 1 february  \textbf{1958  -rrb-  is  an  \underline{australian -- british}} academic cardiologist who has particular interest in the management of heart failure . his research turned established teaching on its head and promoted exercise training -lrb- rather than bed rest -rrb- as a treatment for chronic heart failure . he was instrumental in describing the `` muscle hypothesis '' of heart failure .\\
    \cline{2-4}
     & $Q$: position &$Q$: discipline &$Q$: nationality \\
    \cline{2-4}
     & $A'$: halfback quarterback &$A'$: basketball player& $A'$: philippines filipino\\
    \hline
   
    \multirow{5}{*}{Seq2Seq}&$\tilde D'$: george evans -lrb- born 13 december 1994 -rrb- is an english footballer who plays as a midfielder or centre-back for manchester city .  he was a \textbf{quarterback halfback }in the manchester . & andrei UNK -lrb- born 1975 in satu mare  romania  is a retired romanian aerobic gymnast \textbf{basketball }he had a successful career winning four world championships medals -lrb- two gold , one silver , and one bronze , after his retirement in 1997 he went with to germany where he works as a gymnastics coach at the UNK  \textbf{basketball} club 
  &
   andrew justin stewart coats -lrb- born 1 february 1958  is an \textbf{filipino} -- british academic cardiologist who has particular interest in the management of heart failure . his research turned established teaching on its head and promoted exercise training -lrb- rather than bed rest -rrb- as a treatment for chronic heart failure . he was instrumental in describing the `` muscle hypothesis '' of \textbf{philippines}
\\
    \hline
    
    \multirow{5}{2cm}{With gold  template (SMG(g))} &$\tilde D'$: george evans -lrb- born 13 december 1994 -rrb- is an  english footballer who plays as a \textbf{halfback and quarterback  } for manchester city . &$\tilde D'$: andrei UNK -lrb- born 1975 in satu mare , romania -rrb- is a retired romanian  \textbf{basketball player} . he had a successful career winning four world championships medals -lrb- two gold , one silver , and one bronze -rrb- after his retirement in 1997 he went with to germany where he works as a  \textbf{basketball coach} at the UNK  \textbf{basketball club} in hanover . &
    andrew justin stewart coats -lrb- born 1 february  1958  -rrb-  is  an  \textbf{filipino} academic cardiologist who has particular interest in the management of heart failure . his research turned established teaching on its head and promoted exercise training -lrb- rather than bed rest -rrb- as a treatment for chronic heart failure . he was instrumental in describing the `` muscle hypothesis '' of heart failure .\\
    \hline
     \multirow{5}{2cm}{With predicted  template (SMG(p))}& $\tilde D'$: george evans -lrb- born 13 december 1994 -rrb- is an  \textbf{halfback} footballer who plays as a midfielder \textbf{ or quarterback  } for manchester city .  & andrei UNK -lrb- born 1975 in satu mare , romania  -rrb-  is  a retired romanian \textbf{basketball player} . he had a successful career winning four world championships medals -lrb- two gold , one silver , and one bronze -rrb- after his retirement in 1997 he went with to germany where he works as a  gymnastics coach at the UNK  gymnastic club in hanover . 
     &  $\tilde D'$: andrew justin stewart coats -lrb- born 1 february \textbf{philippines}  academic cardiologist who has particular interest in the management of heart failure . his research turned established teaching on its head and promoted exercise training -lrb- rather than bed rest -rrb- as a treatment for chronic heart failure . he was instrumental in describing the `` muscle hypothesis '' of heart failure .\\
    \hline
    \end{tabular}
    }
    \caption{The example generated cases of competing methods. The underlined tokens are gold answer-related tokens. The bold tokens in the ``Input'' row are predicted answer-related tokens. In the other three rows, the bold tokens are the modified tokens that are related to the given new answer $A'$.}
    \label{tab:showcase}
\end{table*}

Also, in our method, the selection of answer-related words is very important, so we have two evaluations for the selection part: 

\smallskip 
\noindent (1) BLEU (predicted  template) is the BLEU score between the predicted  template (the token sequence after we masked out the answer-related words from the text $D$) and the gold  template (the common sequence of $D$ and $D^\prime$). 

\smallskip 
\noindent (2) Answer $F_1$ measures the Bag-of-words (BOW) $F_1$ value of the generated answer $\tilde A$ compared to the gold answer $A$.  This metric is difficult to achieve, because it requires both to select the correct answer-related tokens and to generate the correct words for the answer $A$.

\subsection{Overall Performance}
We compare our method (SMG) with a baseline method (Seq2Seq). In Seq2Seq, the difference with SMG is that the decoder part is a conventional decoder that completely generates the modified document $\tilde D'$ ignoring the \textit{context  template}. The overall performance is shown in Table~\ref{tab:overall}.

In Table~\ref{tab:overall}, we see that our SMG method has outperformed the Seq2Seq baseline in nearly all evaluation metrics, no matter whether the \textit{context  template} applied to the decoding phase is gold or predicted. Especially, in the two most important metrics for the performance of controllable text edition: iBLEU and diff-BLEU ratio, our model has achieved  a significantly higher score than competing methods. These results show that our method is effective in controllable text edition.

The human evaluation results are also listed in Table~\ref{tab:overall}. The inter-rater agreements are all acceptable ($>0.85$) due to  Krippendorff's principle~(\citeyear{krippendorff2004content}).
According to the human evaluation, when we are using the gold template for partially generating, both the correctness and the fluency of the partially generated text $\tilde D'$ are better than using  the predicted template, which is also consistent with our intuition.  Note that the perplexity score and the fluency score of Seq2Seq are the best of all the three methods; this is because in the partially generated text, the end position of each blank may not fit very well with the next word sometimes, although we have trained an EOA tag. 

Table~\ref{tab:p} shows the experiments  evaluating  the selection of answer-related words. We can see that our SMG model has a higher   BLEU (predicted  template) score than the Seq2Seq model. This fact shows  that partially training the blank-filling tokens helps for the selection of  answer-related tokens. Also, our model SMG has achieved a higher \textit{answer $F_1$} score ($0.68$) than competing methods.

\subsection{Case Study}

We have listed some examples of the modified document $\tilde D'$
generated by the three competing methods (Seq2Seq, SMG(g), and SMG(p)) in Table~\ref{tab:showcase}. We can see that although the answer-related words are already masked out, Seq2Seq  still always generates the words in the original answer $A$ and tends to mix up the words in $A$ and the changed answer $A'$ (like in the second example, Seq2Seq mixed ``gymnastic'' and ``basketball'' together.) Also, Seq2Seq cannot precisely change  everywhere what should be modified, for example, in the second example, Seq2Seq failed to change ``gymnastic coach'' to  ``basketball coach''.   In the SMG methods, when we are using the gold template for partial generation, the model is able to generate the correct words aiming to change $Q$'s answer to $A'$. Although there is still some risk to have some answer-related tokens left unchanged due to the error in the predicted template, the \textit{context tokens} in the predicted template are ensured to be generated. Therefore, our model with predicted template is more fit for NLP products than Seq2Seq.

\section{Conclusion}
In this paper, we proposed a novel task, the goal of  which is to modify some content of a  given text to make the answer of a text-related question change to a given new answer. This task is very useful in many real-world tasks, like contract editing. We constructed a test set for evaluation and released this test set. We also proposed a novel model SMG to solve this task. In SMG, we first use a selector-predictor structure to select the answer-related tokens in the input document, then we use a novel partial generation technique to generate the modified document without changing answer-unrelated tokens in the original document. The experiments proved the effectiveness of our model.

\section*{Acknowledgments}
This work was supported by the EPSRC grant ``Unlocking the Potential of AI for English Law''. We also acknowledge the use of Oxford's Advanced Research Computing (ARC) facility, of the EPSRC-funded Tier 2 facility
JADE (EP/P020275/1), and of GPU computing support by Scan Computers International Ltd.

\bibliography{cite}
\bibliographystyle{acl_natbib}
\appendix

\section*{Appendices}
\section{Human Evaluation Question Marks}
Our annotators were asked the following questions, in order to assess the correctness and fluency of the modified document provided by our model.

\subsection{Correctness of modified document}

Q: Do you think the modification of the document is correct so that it can make the question answer pair $\langle Q, A'\rangle$ true? (For partially correct cases: Partially correct means some places are changed to the new answer, and some places keep the old answer. In this case, only all places (that need to be changed) have been changed can be taken as correct. )

Please choose ``Yes'' or ``No''.

After all human annotators finished their work, the correctness score is calculated by dividing the number of ``Yes'' by the total  number of examples.

\subsection{Fluency}

Q: How fluent do you think the modified document is?  

Please choose a score according to the following description. Note that the score is not necessarily an integer, you can give scores like $3.2$ or $4.9$ , if you deem appropriate.
\begin{itemize}
\item 5: Very fluent. 
\item 4: Highly fluent. 
\item 3: Partial fluent. 
\item 2: Very unfluent. 
\item 1: Nonsense.
\end{itemize}

\section{Experiment Details}

The word embedding size is $300$. 
The BiLSTM  in the selector model has the following hyperparameters: hidden size $= 200$. The hidden size of decoder's LSTM cell is $200$. The rest hyperparameters has the following values: $\lambda_r=1.0$,  $\lambda_\text{eoa}=10$. The hyperparameters are obtained by grid search, the search scopes are  $\lambda_r\in[0.0, 2.0]$ with step size $0.2$, $\lambda_\text{eoa}\in[1, 20]$ with step size $1$, the hidden size are searched in $[100,500]$ with step size $50$. The best hyperparameters are selected when the model achieves the highest answer's $F_1$ in the development set. 
The total parameter size is $72M$.
Each training epoch costs about 1.5 hours on V100 GPU.

\end{document}